\documentclass[conference]{IEEEtran}  
\IEEEoverridecommandlockouts
\usepackage{cite}
\usepackage{amsmath,amssymb,amsfonts}
\usepackage{graphicx}
\usepackage{textcomp}
\usepackage{xcolor}
\usepackage[scientific-notation=true]{siunitx}

\usepackage{algorithm}
\usepackage{algpseudocode}
\usepackage{type1cm}

\algnewcommand{\algorithmicforeach}{\textbf{for each}}
\algdef{SE}[FOR]{ForEach}{EndForEach}[1]
  {\algorithmicforeach\ #1\ \algorithmicdo}
  {\algorithmicend\ \algorithmicforeach}

\newcommand{\Lagr}{\mathcal{L}}
\newcommand{\diff}{\mathrm{d}}

\newcommand{\boldz}{\mbox{\boldmath $z$}}
\newcommand{\boldb}{\mbox{\boldmath $b$}}

\def \figref  #1{\figurename~\ref{#1}}
\def \tabref  #1{\tablename~\ref{#1}}
\def \equref  #1{equation~(\ref{#1})}
\def\BibTeX{{\rm B\kern-.05em{\sc i\kern-.025em b}\kern-.08em
    T\kern-.1667em\lower.7ex\hbox{E}\kern-.125emX}}

\begin{document}

\title{Disentangling Patterns and Transformations \\ from One Sequence of Images \\ with Shape-invariant Lie Group Transformer
}

\author{\IEEEauthorblockN{Takumi Takada\IEEEauthorrefmark{1}, Wataru Shimaya\IEEEauthorrefmark{1}, Yoshiyuki Ohmura\IEEEauthorrefmark{1} and Yasuo Kuniyoshi\IEEEauthorrefmark{1}\IEEEauthorrefmark{2}}
  \IEEEauthorblockA{\IEEEauthorrefmark{1}\textit{School of Information Science and Technology, The University of Tokyo}}
  \IEEEauthorblockA{\IEEEauthorrefmark{2}\textit{Next Generation Artificial Intelligence Research Center (AI Center), The University of Tokyo} \\ Tokyo, Japan\\Email: \{takada, shimaya, ohmura, kuniyosh\}@isi.imi.i.u-tokyo.ac.jp}}

\maketitle

\begin{abstract}
  An effective way to model the complex real world is to view the world as a composition of basic components of objects and transformations.
  Although humans through development understand the compositionality of the real world, it is extremely difficult to equip robots with such a learning mechanism.
  In recent years, there has been significant research on autonomously learning representations of the world using the deep learning; however, most studies have taken a statistical approach, which requires a large number of training data.
  Contrary to such existing methods, we take a novel algebraic approach for representation learning based on a simpler and more intuitive formulation that the observed world is the combination of multiple independent patterns and transformations that are invariant to the shape of patterns.
  Since the shape of patterns can be viewed as the invariant features against symmetric transformations such as translation or rotation, we can expect that the patterns can naturally be extracted by expressing transformations with symmetric Lie group transformers and attempting to reconstruct the scene with them.
  Based on this idea, we propose a model that disentangles the scenes into the minimum number of basic components of patterns and Lie transformations from only one sequence of images, by introducing the learnable shape-invariant Lie group transformers as transformation components.
  Experiments show that given one sequence of images in which two objects are moving independently, the proposed model can discover the hidden distinct objects and multiple shape-invariant transformations that constitute the scenes.
\end{abstract}


\section{Introduction}
When we see an apple fall from a tree, we do not view the scene as a change of an array of thousands of colored pixels.
We derive the semantic components from raw sensory input and decompose the scene into an object and its falling motion. Furthermore, the apple can be subdivided into the leaf and the body, and the motion can be subdivided into its downward movement and rotation.
In this way, viewing the world as a composition of basic components of objects and their changes is an efficient way to model a complex world, and understanding the structure of the world makes it easy to predict its future scene.
Furthermore, by combining a finite number of objects and transformations, we can imagine and simulate an almost infinite number of scenes that cannot possibly exist.

Whereas adults can easily recognize objects and transformations, some research suggests that infants lack the ability to do so.
It has been experimentally shown that infants are immature in understanding the geometric shape of an object and are unable to achieve a representation of its abstract shape\cite{smith2009fragments}.
In addition, it has been shown that infants under 18 months old had difficulty solving the shape-sorter toys\cite{ornkloo2007fitting}, which implies that infants cannot discover the rotational transformation applied to the shaped blocks and recognize the shapes of holes as being different from those of the blocks.
These studies suggest that the ability to recognize objects and transformations is acquired through development.

Models that attempt to autonomously learn the compositionality of observed images are called representation learning models, and have been studied extensively.
Recently, some generative models using deep neural networks have been proposed\cite{higgins2017beta,chen2016infogan} and they are said to discover underlying generation factors (e.g., position, size, and so forth) in an unsupervised manner from raw images.
Some models can also handle multi-object scenes through the iterative application of a generative model on an image\cite{burgess2019monet,greff2019multi}.
However, in such probabilistic approaches, information about objects and transformations is jointly represented in the latent space.
In other words, models do not consider objects and transformations as separate things as humans do.
In addition, probabilistic approaches require a huge number of training data.
The ability to process unknown objects is important to realize generalizable recognition and prediction, similar to those of humans.
If attempting to realize transformations applicable to any objects with statistical models, the number of training data has to infinitely increase.
Furthermore, it has been theoretically and empirically proven that it is impossible to learn a disentangled representation based solely on statistical independence\cite{locatello2019challenging}.

Instead of statistical properties, it is natural to make use of algebraic properties.
Otsu proposed a pattern recognition theory using Lie group theory\cite{otsu1986recognition}.
According to the theory, information that patterns contain is separated into two kinds of features, features invariant to transformations and the ones affected by transformations.
For example, even if a certain pattern is moved or rotated, still we can recognize the pattern because those symmetry transformations affect the position or the angle but do not affect the information about the identity of patterns such as the shape.
Some models were developed to obtain the symmetry transformations from the images by trying to approximate the Lie group operator by the spatio-temporal matrix filters\cite{miao2007learning,olshausen2007bilinear,DBLP:journals/corr/abs-1001-1027,memisevic2010learning,chau2020disentangling}.
However, those models require a lot of training data, usually more than 1000, because they estimate parameters using stochastic methods. 
Regarding this problem, Takada~\textit{et al.} focused on the fact that transformations defined by ordinary differential equations (ODE) always satisfy requirements of the Lie group and proposed a model to discover a shape-invariant symmetry transformation from few examples by exploiting the a priori embedding of Lie group properties\cite{takada2021unsupervised}. This method is, however, not equipped with pattern-identification mechanism.
According to the theory, the identity of patterns can be viewed as the invariant features against symmetry transformations; therefore, we can expect that the patterns can naturally be extracted by expressing transformations with symmetric Lie group transformers and reconstructing the scenes with them.
Based on this idea, we propose a novel approach to disentangle the scenes into the minimum number of both patterns and transformations without any supervision, by introducing the shape-invariant Lie group transformer\cite{takada2021unsupervised} to represent symmetry transformations.
In experiments, we show our proposed method can discover the hidden components of patterns and transformations that constitute the scenes only from one sequence of images.

\section{Formulation}
We first describe how the recognition of patterns and transformations is formulated in the research by Otsu\cite{otsu1986recognition}.
In this theory, patterns can be described as the function $p(x,y) \ (x,y \in \mathbb{R})$ and the set of such functions forms the pattern space $P_2$. Here, a transformation applied to the pattern is a mapping $T$ from $P_2$ to $P_2$.
A certain shape-invariant transformation to a pattern $p$ generally consists of a combination of several basic shape-invariant transformations with one transformation quantity parameter $\lambda \in \mathbb{R}$, which as a whole form a continuous group.
\begin{equation}
  \label{equ:basic_trs}
  T(\lambda_1, \dotsc, \lambda_N) p = T_N(\lambda_N)\cdots T_1(\lambda_1) p.
\end{equation}
Each basic transformation $T_k(\lambda_k)$ is assumed as a Lie group operator, which meets the following conditions:
\begin{align}
  T(\mu) + T(\lambda) & = T(\lambda + \mu) \notag                                                     \\
                      & = T(\lambda) + T(\mu) \  (\lambda, \mu \in \mathbb{R}) \label{equ:additive}   \\ 
  T(0)                & = I \hspace{4pt} \hspace{8pt} \text{(Identity operator)} \label{equ:identity} \\ 
  {T(\lambda)}^{-1}   & = T(-\lambda) \hspace{8pt} \text{(Inverse operator)}. \label{equ:inverse}     
\end{align}

The recognition of pattern $p$ can be formulated as obtaining a function $\Psi$ which returns the same value for the same patterns, irrespective of what transformation it is applied to. Therefore, the function $\Psi$ meets the equation: $0 = \Psi[T(\lambda) p] - \Psi[p]$.
By contrast, recognizing the transformation $T_k$ is to obtain a function $\Phi$ that derives the corresponding transformation quantity $\lambda_k$ from the patterns before and after the transformation, as is in the equation: $\lambda_k = \Phi[T_k(\lambda_k) p] - \Phi[p]$.

\begin{figure}[htbp]
  \centering
  \includegraphics[width=0.45\textwidth]{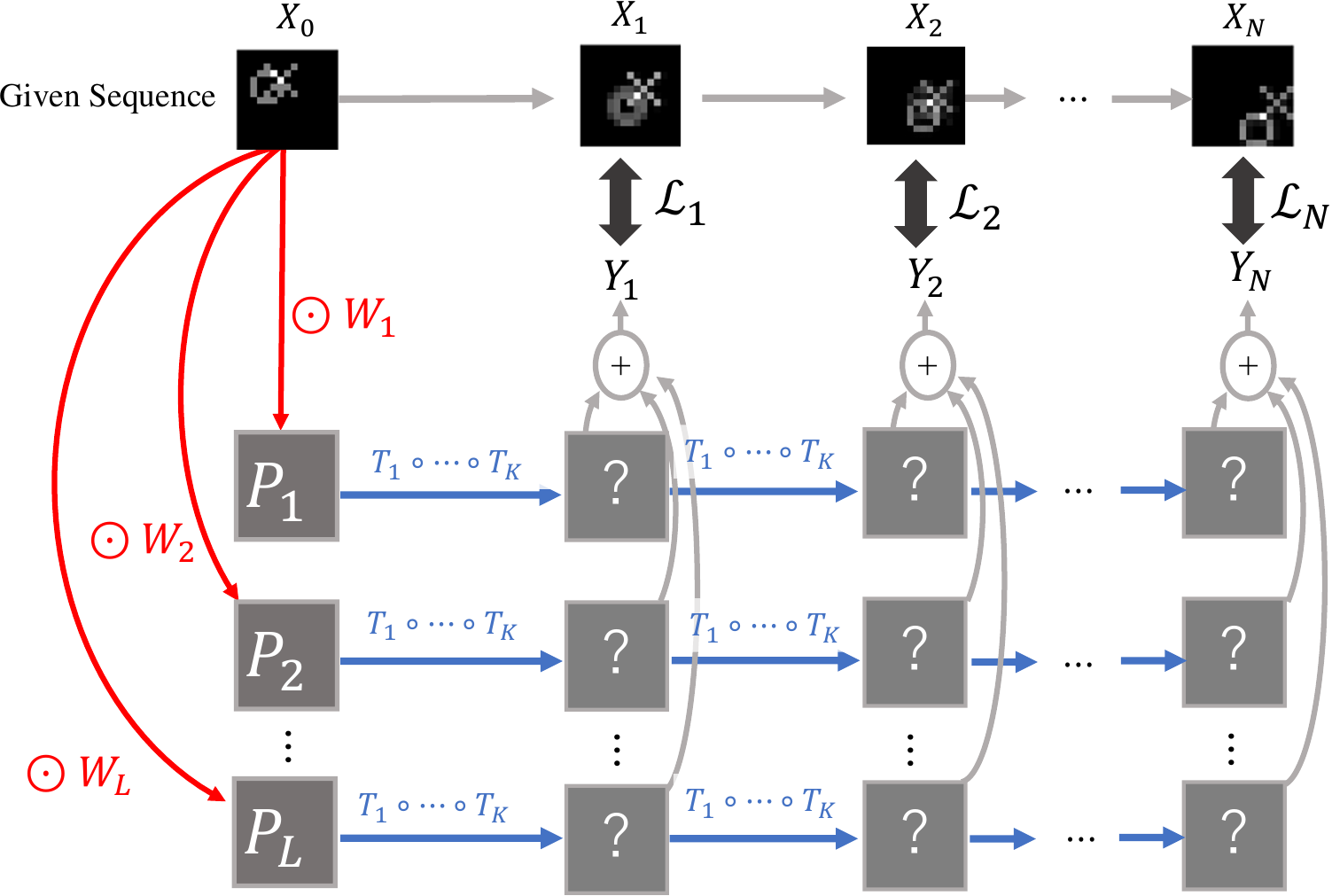}
  \caption{Architecture of composition learning model. All patterns $P_l$, transformers $T_k$ and variables $\lambda_{k,l,i}$ are learning targets.}
  \label{fig:arch_composition_learning}
\end{figure}

Based on the above, if we assume that the observed sequential images of the changing world $\{X_i | i = 0, 1, \dotsc , N\}$ are expressed as a superimposition of multiple patterns independently undergoing multiple transformations, the observed image $X_i$ at timestep $i$ in the sequence can be described as follows:
\begin{align}
  \label{equ:compositionality}
  X_i & = \sum_{l=1}^L \left( \prod_{k=1}^K T_k (\lambda_{k, l, i}) \right) P_l                                         \\
      & = T_1(\lambda_{1,1, i}) \circ T_2(\lambda_{2, 1, i}) \circ \dots \circ T_K(\lambda_{K, 1, i}) P_1 \notag        \\
      & \quad + T_1(\lambda_{1, 2, i}) \circ T_2(\lambda_{2, 2, i}) \circ \dots \circ T_K(\lambda_{K, 2, i}) P_2 \notag \\
      & \quad + \dotsb \notag                                                                                           \\
      & \quad + T_1(\lambda_{1,L, i}) \circ T_2(\lambda_{2, L, i}) \circ \dots \circ T_K(\lambda_{K, L, i}) P_L. \notag
\end{align}
Note that $P_1, P_2, \dotsc , P_L \ (L \in \mathbb{N})$ are distinct patterns that exist in the sequence, $T_1, T_2, \dotsc , T_K \ (K \in \mathbb{N})$ are the basic shape-invariant transformations with one parameter, and $\lambda_{k,l,i}$ is a transformation quantity of transformation $T_k$ that is applied to the pattern $P_l$ in the scene $X_i$ from the initial scene $X_0$.
In this formulation, we define pattern primitives $P_1,\dotsc,P_L$ using the initial scene $X_0$. Meaning, the following equation stands:
\begin{equation}
  \label{eq:sum_pi_x0}
  P_1+P_2+\cdots +P_L=X_0.
\end{equation}
We use the product operator $\prod T_k$ to denote the composite functions $T_1\circ T_2\circ \cdots \circ T_K$.
\figref{fig:arch_composition_learning} shows how the scene $X_i$ should be reconstructed with patterns $P_1,\dotsc,P_L$ and transformations $T_1,\dotsc,T_K$.
We aim to develop a model that can determine those hidden basic patterns $P_1, \dotsc, P_L$ and transformations $T_1, \dotsc, T_K$ from only one observed sequence without supervision.

\section{Proposed Model}
\subsection{Model architecture}
Let us suppose a sequence of images $X_\mathrm{seq} = \{X_i | i=0, 1, \dotsc, N \}$ in which several transformations are applied to objects independently. Note that in this study, for simplicity, we assume that the observed images are gray-scale images in which the regions where the patterns are placed are activated and pixel values of the background are set to zero.
The goal of our proposed model is to determine the patterns and transformers by which the given sequence can be reconstructed.
To do so, we initialize the multiple untrained pattern primitives and transformers with which we build architecture as is shown in \figref{fig:arch_composition_learning}. We then update the parameters of patterns and transformers so that the error between $X_\mathrm{seq}$ and reconstructed sequence $Y_\mathrm{seq} = \{Y_i | i = 0,1,\dotsc,N \}$ is minimized. 
Here, reconstructed scene $Y_i$ is computed as $Y_i = \sum_{l=1}^L \left( \prod_{k=1}^K \tilde{T}_k (\tilde{\lambda}_{k, l, i}) \right) \tilde{P}_l$ and  $\tilde{P}$ and $\tilde{\lambda}$ represent estimated pattern primitives and transformation quantities respectively, and $\tilde{T}$ represents the transformation functions whose model parameters are estimated parameters. In the following paragraph, we describe the detailed descriptions of pattern primitives and transformers.

Pattern primitives are initialized as images $P_l \in \mathbb{R}^{H\times W}$, where $H$ and $W$ are the height and width of each image.
As mentioned in the \equref{eq:sum_pi_x0}, the superimposition of pattern primitives should be equal to the initial scene $X_0$, we thus obtain $P_l$ by applying weight matrix $W_l$ with the initial image $X_0$ (i.e., $P_l = X_0 \odot W_l$) and guarantee that the sum of the patterns becomes $X_0$.
Such weight matrices are initialized by a random number and updated through training.

As for transformers, we employ the shape-invariant Lie group transformer\cite{takada2021unsupervised}, which is embedded with an ODE in its dynamics.
In practice, this dynamics is implemented by NeuralODE\cite{chen2018neural} and applies a transformation by moving pixels individually to another location to guarantee the invariance to the shape of the input pattern.
The destination of each pixel $[x(t), y(t)]^T$ is determined by solving an ODE using the transformation quantity $\lambda$ as the time for the ODE and the initial position of the pixel as the initial value, as is in the following \equref{eq:ode_func}:
\begin{equation}
  \label{eq:ode_func}
  \frac{\diff}{\diff t} \begin{bmatrix} x(t) \\ y(t) \end{bmatrix}
  = f \left(
  \begin{bmatrix} x(t) \\ y(t) \end{bmatrix}
  \right)
  = A \begin{bmatrix} x(t) \\ y(t) \end{bmatrix}
  + \boldb .
\end{equation}
Here, the coordinates are mapped so that the center, bottom-right, and top-left points in the image correspond to $[0,0], [1,1], [-1,-1]$, respectively.
The parameters $A \in \mathbb{R}^{2\times2}$ and $\boldb \in \mathbb{R}^2$ are the model parameters that are initialized with random numbers and updated through training.
Hereinafter, these parameters $A$ and $\boldb$ are collectively referred to as $\theta$.
The main features of this transformer are (a) satisfying the properties (\equref{equ:additive}(\ref{equ:identity})(\ref{equ:inverse})) of the Lie group operators, and (b) the applicability of a transformation to any unknown shapes of patterns.
Especially the feature (b) is important, because if the transformer is variant to different patterns, we cannot combine the transformation and pattern components freely and express new scenes, which means we would need a new transformer for each pattern and this is extremely inefficient for modeling the world.

In addition, transformation quantities $\Delta \lambda_{k,l,i}$ are randomly initialized for each transformation, pattern, and timestep. Note that the value $\Delta \lambda_{k,l,i}$ denotes the transformation quantity of transformation $T_k$ applied to pattern $P_l$ at scene $X_i$ from the previous scene $X_{i-1}$ and the variable $\lambda_{k,l,i}$ is expressed as the following equation:$\lambda_{k,l,i} = \sum_{i'=1}^{i} \Delta \lambda_{k,l,i'} \ \  (\lambda_{k,l,0} = 0)$.

Because the model does not know how many patterns and transformations are hidden in the given sequence prior to training, we have to prepare a redundant number of untrained pattern primitives and transformers, and throughout the training, we expect such redundant components to be identity elements. In other words, we expect redundant pattern primitives to be zero matrices, and redundant transformers to be identity mappings.

\subsection{Objective function}
Using initialized patterns, transformers and transformation quantities, the estimated scene $Y_i$ can be obtained by superimposing the estimated patterns that several estimated transformations are applied to.
All estimated patterns, model parameters of transformers and transformation quantities are optimized so that the error between the observed sequence and the reconstructed sequence is minimized.
Now, we consider the objective function $\Lagr_P$ for the learning patterns and $\Lagr_T$ for the transformers.
With such objective functions, we aim to obtain the optimal parameters $\hat{P}, \hat{\theta}$, and $\hat{\lambda}$ by solving the equations~(\ref{eq:objectives}) below:
\begin{equation}
  \hat{P} = \min_P \Lagr_P(X_\mathrm{seq}, Y_\mathrm{seq}, P) \hspace{10pt}
  \hat{\theta}, \hat{\lambda} = \min_{\theta, \lambda} \Lagr_T(X_\mathrm{seq}, Y_\mathrm{seq}, \theta, \lambda). \\
  \label{eq:objectives}
\end{equation}
In our proposed method, we aim to obtain such optimal patterns and transformations using the gradient descent, and the algorithm for obtaining them is shown in Algorithm \ref{alg1}.
\begin{figure}[htbp]
  \small
  \begin{algorithm}[H]
    \caption{Composition Learning Algorithm}
    \label{alg1}
    \begin{algorithmic}[1]    
      \State Initialize $\{P, \theta, \lambda \} $ \algorithmiccomment{Initialization}
      \While{$\{P, \theta, \lambda \}$ not converged}

      \ForEach{$i \in \{0, 1, \ldots, N \}$}
      \State $Y_i \leftarrow \sum_{l=1}^L \left( \prod_{k=1}^K \tilde{T}_k (\tilde{\lambda}_{k, l, i}) \right) \tilde{P}_l$  
      \EndForEach

      \State $\Delta \theta \leftarrow \nabla_{\theta} \Lagr_T$
      \State $\theta \leftarrow \mathrm{update}(\theta, \Delta \theta)$ \algorithmiccomment{Update $\theta$}
      \State $\Delta \lambda \leftarrow \nabla_{\lambda} \Lagr_T$
      \State $\lambda \leftarrow \mathrm{update}(\lambda, \Delta \lambda)$  \algorithmiccomment{Update $\lambda$}

      \ForEach{$i \in \{0, 1, \ldots, N \}$}
      \State $Y_i \leftarrow \sum_{l=1}^L \left( \prod_{k=1}^K \tilde{T}_k (\tilde{\lambda}_{k, l, i}) \right) \tilde{P}_l$ 
      \EndForEach

      \State $\Delta P \leftarrow \nabla_{P} \Lagr_P$
      \State $P \leftarrow \mathrm{update}(P, \Delta P)$  \algorithmiccomment{Update $P$}

      \EndWhile

    \end{algorithmic}
  \end{algorithm}
\end{figure}

Patterns and transformers are learned to better reconstruct the given sequence.
We use the mean square error (MSE) loss as the reconstruction loss for pattern training. By contrast, we employ the masked-MSE\cite{takada2021unsupervised} as the reconstruction loss for a training transformation.
In addition to the reconstruction losses, auxiliary loss functions are introduced to the full objective. These loss functions are additional constraints so that the model can reconstruct the given sequence only with the minimum number of components.

The auxiliary loss function for learning patterns is the pattern-entropy loss $\Lagr_\mathrm{ptn-entropy}$, as is defined in \equref{eq:ptn-entropy}.
\begin{equation}
  \label{eq:ptn-entropy}
  \Lagr_\mathrm{ptn-entropy} = - \sum_{l=1}^L Q_l\log Q_l.
\end{equation}
Intuitively, patterns that make the exactly same movements should be regarded as the same pattern.
To compute the pattern-entropy $\Lagr_\mathrm{ptn-entropy}$, the value $Q_l$, which is the ratio of the area occupied by an estimated pattern $P_l$ in the observed image $X_0$, is computed and we compute this ratio for all estimated patterns and obtain the average amount of information (entropy) by viewing these ratios as probabilities.
By minimizing this entropy, we aim to encourage the model to lump all patterns that are moving together.
Therefore, the full objective function for the learning patterns is \equref{eq:lagr_p}:
\begin{equation}
  \label{eq:lagr_p}
  \Lagr_P = \sum_{i=1}^N r^i ||X_i - Y_i||_2^2 + \alpha \Lagr_\mathrm{ptn-entropy} \hspace{8pt} (r, \alpha \in \mathbb{R}).
\end{equation}
Here, the variable $r$ is a discount rate such that the reconstruction in the near future will be prioritized, and the variable $\alpha$ is the coefficient for the auxiliary loss.

To train the transformers, we introduce three auxiliary loss functions. The first auxiliary loss $\Lagr_\mathrm{L1-reg}$ is L1-regularization for the transformers. This encourages the transformers to have simpler parameters.
Now that we are applying the L1 regularization to the model parameters $A$ and $\boldb$, the scale of transformation quantity $\lambda$ should be fixed because otherwise the scale of the model parameters (the values in $A$ and $\boldb$) can be infinitely small. We thus introduce $\lambda$-scale-fixing loss $\Lagr_\mathrm{\lambda-scale}$.
Using method, we expect the largest estimated transformation quantity to be 1.
Thus, $\Lagr_\mathrm{\lambda-scale}$ is computed as the sum of errors between estimated transformation quantities $\tilde{\lambda}_{k,l,i}$ and normalized transformation quantities $\frac{\tilde{\lambda}_{k,l,i}}{ \max{\{\lambda_{k,l,N} | l=1,2,\dotsc,L \} }}$.
Furthermore, we introduce the inner product loss between any two transformers. Generally speaking, the conditions for the value of the inner product to be zero are either two vectors becoming orthogonal, or the norm of either one of the vectors becoming zero. For the latter reason, we can expect a redundant transformer to be the identity mapping.
These losses are defined in the following equations (\ref{eq:l1-reg})(\ref{eq:lambda-scale})(\ref{eq:inner-prod}):
\begin{align}
  \Lagr_\mathrm{L1-reg}         & = \sum_{k=1}^K ||\tilde{\theta}_k||_1                                                                                              \label{eq:l1-reg} \\
  \Lagr_\mathrm{\lambda -scale} & = \sum_{k,l,i} \left\| \tilde{\lambda}_{k,l,i} - \frac{\tilde{\lambda}_{k,l,i}}{\max_{l\in \{1,...,L\}}{\{\tilde{\lambda}_{k,l,N}\}}} \right\|       \label{eq:lambda-scale}
\end{align}
\begin{align}
  \Lagr_\mathrm{inner-prod} & = \sum_{i=1}^{K-1} \sum_{j=i+1}^K \langle \tilde{A}_i, \tilde{A}_j, \rangle + \langle \tilde{\boldb}_i, \tilde{\boldb}_j \rangle. \label{eq:inner-prod}
\end{align}
Here, $\tilde{A}_k, \tilde{\boldb}_k \ (k=1,2,\dotsc,K)$ refer to the model parameters of the transformer $T_k$, and we compute the inner product between matrices by vectorizing them.
With these loss functions, the full objective for learning the transformers and transformation quantities is the following \equref{eq:lagr_t}:
\begin{align}
  \label{eq:lagr_t}
  \Lagr_T = \sum_{i=1}^N r^i ||X_i - Y_i\odot X_i||_2^2 + \beta\Lagr_\mathrm{L1-reg} \notag \\
  + \gamma \Lagr_\mathrm{\lambda-scale} + \delta \Lagr_\mathrm{inner-prod} \hspace{6pt} (\beta, \gamma, \delta \in \mathbb{R}).
\end{align}

\section{Experiments}
\subsection{Datasets and Experimental Setting}
\begin{figure}[htbp]
  \centering
  \includegraphics[width=0.4\textwidth]{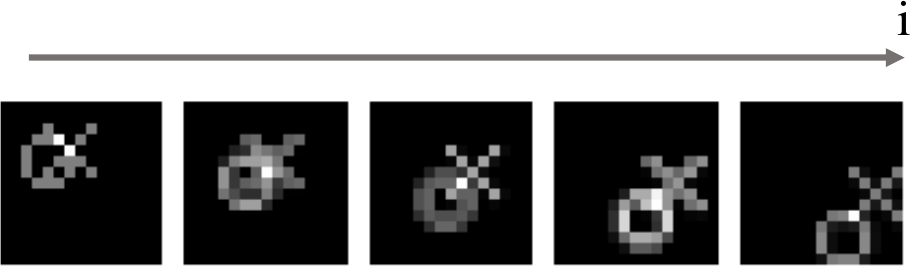}
  \caption{The sequence of images ($15\times15$ in size) given to the model. Two objects ``X'' and ``O'' are translated independently.}
  \label{fig:dataset}
\end{figure}

Using the above proposed model, we conducted an experiment to determine distinct basic hidden patterns and transformations from a single sequence.
We trained our model on our custom-dataset in which ``X''-shaped and ``O''-shaped patterns are independently moving in the XY-plane. The sequence of images used in this experiment is shown in \figref{fig:dataset}.
The goal of this experiment is to obtain these two distinct patterns and two mutually orthogonal transformations such that translations to any directions can be realized.
In this experiment, because the given sequence contains only two patterns and transformations, we redundantly prepare three pattern primitives and transformers. We set the coefficient values for the auxiliary loss $\alpha, \beta, \gamma, \delta$ as $\num{0.001}, \num{0.0001}, \num{0.1}, \num{0.0001}$, respectively.

\subsection{Results}  
\begin{figure}[htbp]
  \centering
  \includegraphics[width=0.45\textwidth]{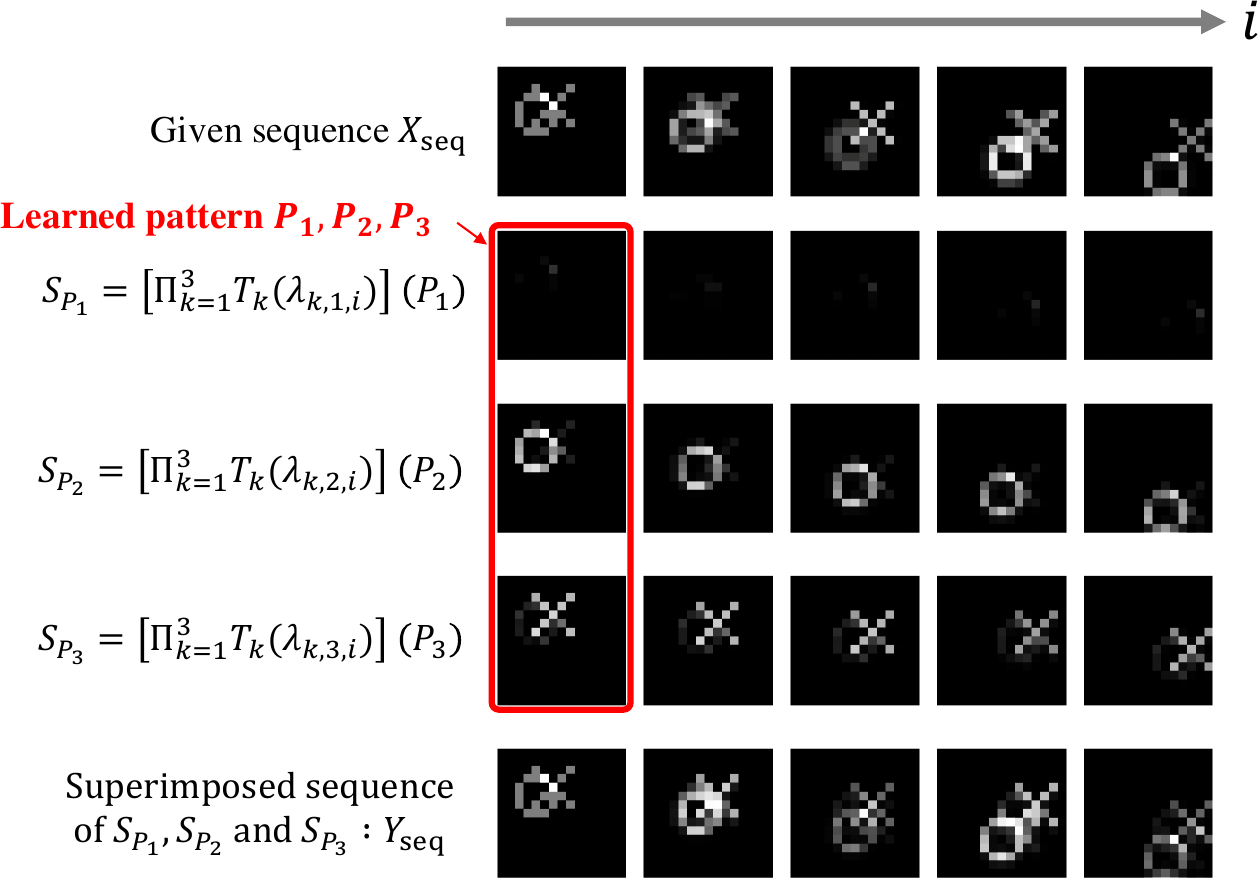}
  \caption{The fist row is the given sequence. Three patterns $P_1, P_2$, and $P_3$ in the red box are learned patterns. The second, third and fourth rows at the top show the estimated future sequence for each pattern, and the bottom row is the reconstructed sequence. The images are thresholded for a clear view.}
  \label{fig:result_ptnn_trsk}
\end{figure}

After training, we found that distinct patterns and transformations that constitute the given sequence were obtained without any supervision.
Obtained pattern primitives and the reconstructed sequence are shown in \figref{fig:result_ptnn_trsk}.
The top row is the given sequence, and the images surrounded by the red line are obtained pattern primitives $P_1, P_2$, and $P_3$.
The second, third and fourth rows at the top are the sequences $S_{P_1}, S_{P_2}$, and $S_{P_3}$ generated by applying the composite transformation $T_3 \circ T_2\circ T_1$ to $P_1, P_2, P_3$ respectively. The sequence at the bottom row $Y_\mathrm{seq}$ is the reconstructed sequence generated by superimposing sequences $S_{P_1}, S_{P_2}$ and $S_{P_3}$.
From this figure, we can observe that the sequence is reconstructed very well.
In addition, we can see that two distinct patterns (``O''-shaped pattern as $P_2$ and ``X''-shaped pattern as $P_3$) are discovered, and the third primitive, which was redundant, converged at zero (black image).

\begin{figure}[htbp]
  \centering
  \includegraphics[width=0.48\textwidth]{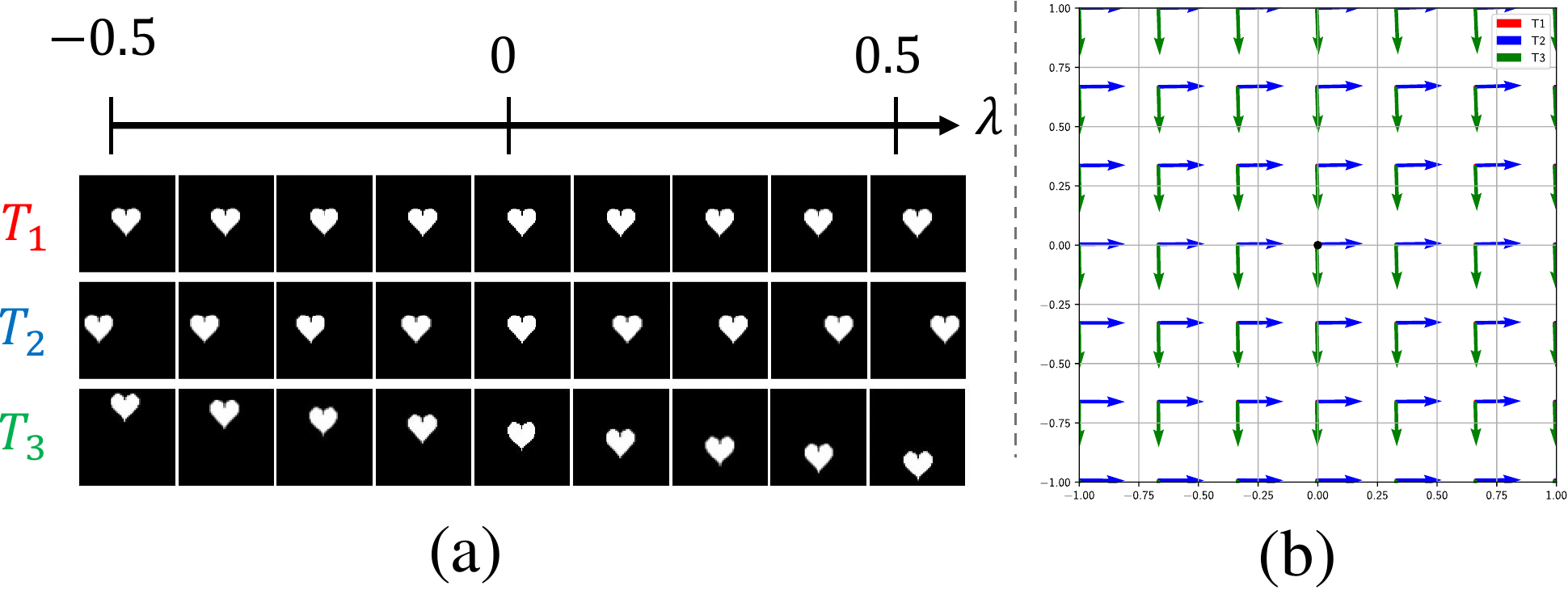}
  \caption{(a) The transitions of a pattern undergoing transformations from $T_1, T_2$ and $T_3$. (b) Field that each learned transformer forms.}
  \label{fig:traverse_field}
\end{figure}

We also evaluate the obtained transformers $T_1, T_2$ and $T_3$.
To see what type of transformers they converged into, we apply the obtained transformer to an untrained object and observe the transition made by each transformer.
\figref{fig:traverse_field}(a) shows how a pattern changes as it undergoes transformations from each obtained transformer, and the first, second and bottom rows correspond to the transition by the transformer $T_1, T_2$ and $T_3$, respectively. The horizontal axis indicates the amount of transformation applied to the object with the middle point set to zero, and the transformation quantity $\lambda$ increases from left to right.
The images at the center column ($\lambda=0$) are identical to the original images before the transformation. From \figref{fig:traverse_field}(a), we can see that the transformer $T_1$ is an identity map that do not move the pattern and the transformers $T_2$ and $T_3$ are translations that move the pattern horizontally and vertically, respectivelly.

The visualized transformation fields formed by each obtained transformer are shown in \figref{fig:traverse_field}(b). The red, blue and green arrows within the field correspond to the transformers $T_1, T_2$ and $T_3$, respectively. These arrows represents the direction and the magnitude of the transformation applied to each point. Specifically, these arrows represent the gradient vector $A[x, y]^T + \boldb$ at each point $[x, y]^T$. Note that the variables $A$ and $\boldb$ are model parameters mentioned in \equref{eq:ode_func}.
From \figref{fig:traverse_field}(b), we can see that the blue arrows of the field for transformer $T_2$ are pointing toward the right, and the green arrows for the transformer $T_3$ are pointing to the bottom. Therefore, it can be said that the translation transformers $T_2$ and $T_3$ became the mutually orthogonal translation transformers. Furthermore, the arrows formed by the transformer $T_1$, which are in red, can not be observed. This implies that the transformation field of the transformer $T_1$ converged into a zero-vector field.

The exact values of $A$ and $\boldb$ for each transformer obtained are shown in \tabref{tab:params_ptnn_trsk} and relatively large values are highlighted in bold. From this table, we can see that the values of elements in the weight matrix $A$ of all transformers are relatively small and can be ignored. In transformer $T_2$, the first value of the bias term $\boldb$  is large, and in the transformer $T_3$, the second values in the bias term $\boldb$ is large.
Generally speaking, the horizontally or vertically translating point $\boldz(\lambda) = [x(\lambda), y(\lambda)]^T$ conforms to the ODEs
\begin{align}
  \frac{\diff \boldz(\lambda)}{\diff \lambda}
  =  \begin{bmatrix}
       k \\ 0
     \end{bmatrix}, \,
  \frac{\diff \boldz(\lambda)}{\diff \lambda}
  = \begin{bmatrix}
      0 \\ k
    \end{bmatrix} \, (k \in \mathbb{R}),
\end{align}
respectively.
Therefore, it can be deduced that the transformer $T_2$ converged at a horizontal translation and $T_3$ converged at a vertical translation.
Because those two translations are not parallel, it can be deduced that the transformers $T_2$ and $T_3$ are linearly independent. By contrast, the bias term of the transformer $T_1$ is small and near zero, thus we can see that transformer $T_1$ converged at the identity mapping.

\begin{table}[htbp]
  \setlength{\tabcolsep}{2pt}
  \centering

  \caption{ODE parameters for each transformer.}
  \footnotesize
  \begin{tabular}{c}
    \hline
    transformer 1                                   \\ %
    \begin{tabular}{ll}
      \smallskip
      \footnotesize
      $A = \left[
          \begin{array}{rr}
            \num{0.0011}  & \num{-0.016} \\  %
            \num{-0.0036} & \num{0.0063} %
          \end{array}
      \right]$ \normalsize &
      \footnotesize
      $ \ \boldb = \left[
          \begin{array}{r}
            \num{0.00056} \\  %
            \num{0.023}
          \end{array}
          \right]$ \normalsize
    \end{tabular} \\  %
    transformer 2                                   \\ %
    \begin{tabular}{ll}
      \smallskip
      \footnotesize
      $A = \left[
          \begin{array}{rr}
            \num{-0.0049} & \num{0.0095}  \\  %
            \num{-0.0014} & \num{-0.0024} %
          \end{array}
      \right]$ \normalsize &
      \footnotesize
      $ \ \boldb = \left[
          \begin{array}{r}
            \num[math-rm=\mathbf]{0.97} \\  %
            \num{0.014}
          \end{array}
          \right]$ \normalsize
    \end{tabular} \\  %
    transformer 3                                   \\ %
    \begin{tabular}{ll}
      \smallskip
      \footnotesize
      $A = \left[
          \begin{array}{rr}
            \num{0.0018}  & \num{0.017}  \\  %
            \num{-0.0085} & \num{0.0072} %
          \end{array}
      \right]$ \normalsize &
      \footnotesize
      $ \ \boldb = \left[
          \begin{array}{r}
            \num{0.046} \\  %
            \num[math-rm=\mathbf]{1.0}
          \end{array}
          \right]$ \normalsize
    \end{tabular} \\  %

    \hline
  \end{tabular}
  \label{tab:params_ptnn_trsk}
\end{table}

\subsection{Experiments with different sequences}
\begin{figure}[H]
  \centering
  \includegraphics[width=0.4\textwidth]{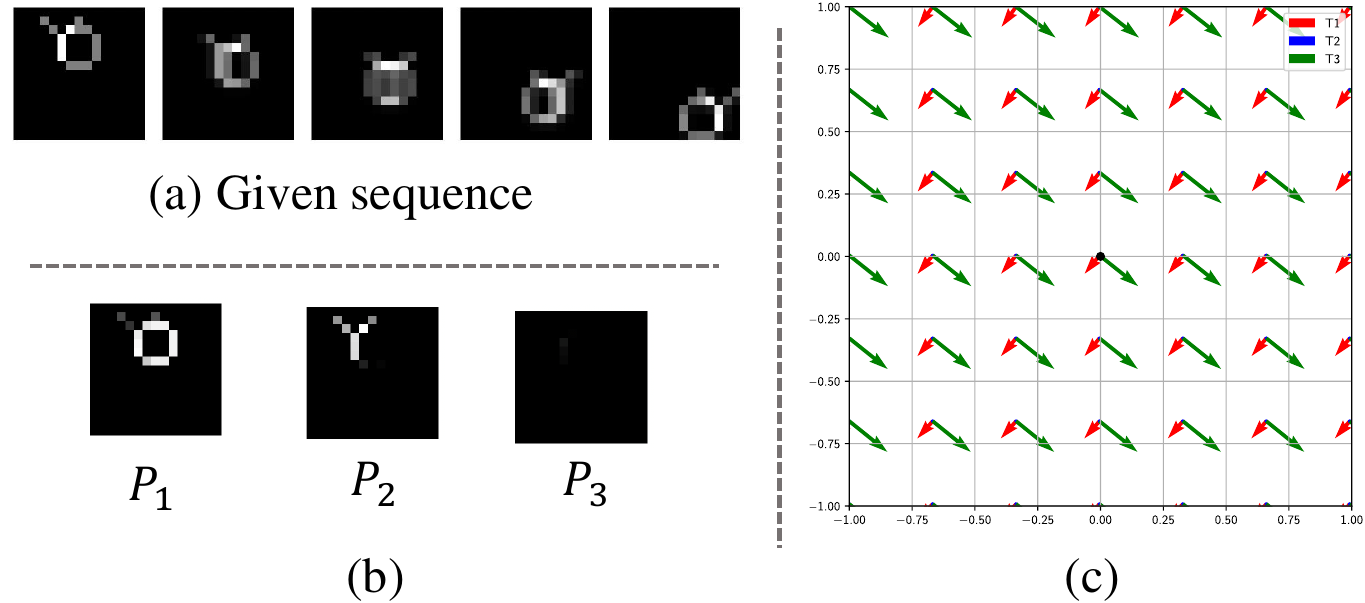}
  \caption{(a) Given sequence, (b) Obtained patterns, (c) Fields of obtained transformers.}
  \label{fig:eg1}
\end{figure}

\begin{figure}[htbp]
  \centering
  \includegraphics[width=0.4\textwidth]{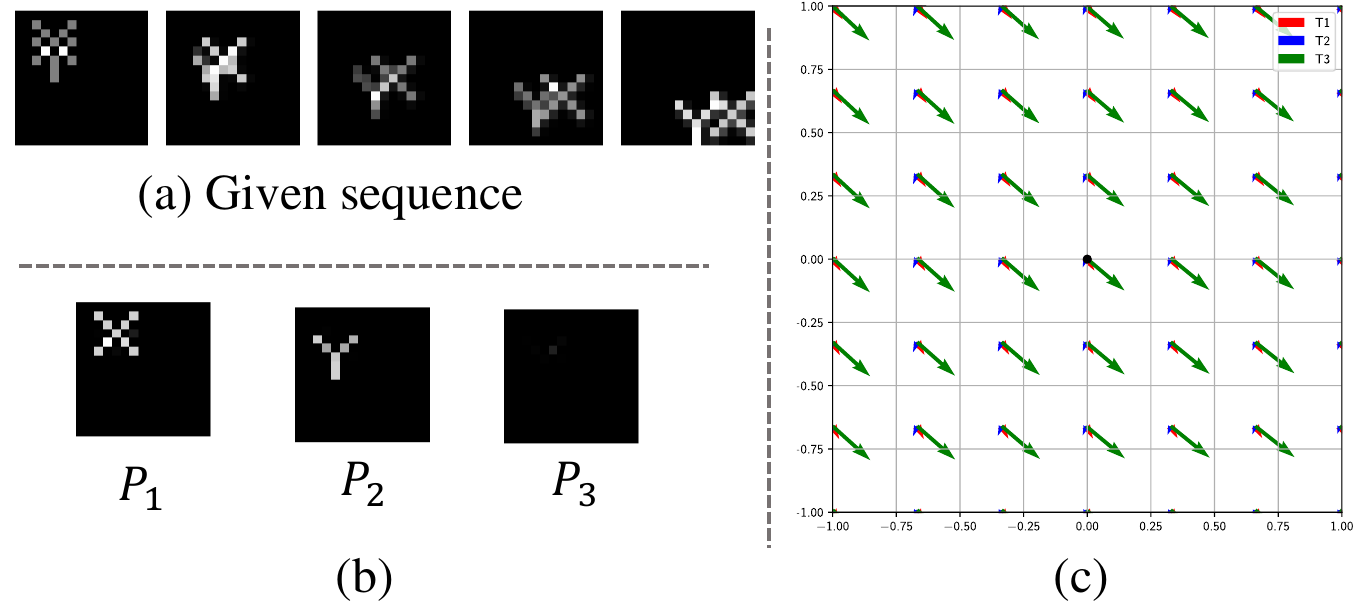}
  \caption{(a) Given sequence, (b) Obtained patterns, (c) Fields of obtained transformers.}
  \label{fig:eg2}
\end{figure}

We conducted the same experiments with different sequences and the results are shown in \figref{fig:eg1} and \figref{fig:eg2}.
In each figure, (a) shows the given sequence, (b) shows obtained patterns and (c) shows the fields formed by the obtained transformers.
From \figref{fig:eg1}, we can see that two patterns and two orthogonal transformers were successfully obtained from the sequence in which ``Y''-shaped and ``O''-shaped patterns were moving.
As for \figref{fig:eg2}, we can see that while two distinct patterns were properly discovered, only one translation transformation was obtained and other two transformers converged to identity mappings. This is because two patterns are actually moving in almost the same direction and only one translation is sufficient to reconstruct the scenes.

\section{Conclusion}
In this study, a novel approach to breaking an observed image down to the minimum number of basic components is proposed.
We hypothesized that the observed world is made of a combination of basic patterns and transformations, and we built architecture to determine the hidden basic components of patterns and transformations.
We expected that the identity of a pattern such as the shape can naturally be extracted by representing the transformations to the pattern as the symmetry transformations and attempting to reconstruct the observed scene with them.
We thus employed the shape-invariant Lie group transformer to represent the symmetry transformations. We trained our model on our custom dataset in which two distinct patterns move independently and showed that the model can acquire such distinct patterns and mutually orthogonal transformers from only one sequence of images, with unnecessary elements converging into identity elements. We trained the multi-object VAE\cite{burgess2019monet} on the same dataset and observed that it failed to discover the hidden patterns and transformations from such a small dataset.
Our model shows some important features of human-like intelligent agents that can discover the components of the scenes from few experiences.
This ability is important because understanding the structure of the world makes it easy to predict future world scenes and once the basic components are obtained, by combining them, nearly an infinite number of new scenes can be simulated.

However, our proposed model has several issues to be resolved.
The Lie group transformer used in our model can deal with the rotation only when the center of rotation is fixed\cite{takada2021unsupervised}.
If a pattern is moving while rotating, the center of rotation will not be fixed.
Therefore, our model cannot currently deal with the translation-rotation disentanglement, which is an urgent issue.
Another issue is handling a sequence in which more complicated transformations such as a deformation occurs.
Our research is based on the research about the shape-invariant transformations, and the free-form deformation, which will likely to be variant to the pattern (e.g. font changes of characters), is outside the scope of the original formulation\cite{otsu1986recognition} thus generalizing the theory would be an important future issue.

\section*{Acknowledgment}
This work was supported by Next Generation Artificial Intelligence Research Center (AI Center), The University of Tokyo and the Donated Chair of Frontier AI Research and Education, School of Information Science and Technology, The University of Tokyo.


\begin{thebibliography}{10}
\providecommand{\url}[1]{#1}
\csname url@samestyle\endcsname
\providecommand{\newblock}{\relax}
\providecommand{\bibinfo}[2]{#2}
\providecommand{\BIBentrySTDinterwordspacing}{\spaceskip=0pt\relax}
\providecommand{\BIBentryALTinterwordstretchfactor}{4}
\providecommand{\BIBentryALTinterwordspacing}{\spaceskip=\fontdimen2\font plus
\BIBentryALTinterwordstretchfactor\fontdimen3\font minus
  \fontdimen4\font\relax}
\providecommand{\BIBforeignlanguage}[2]{{%
\expandafter\ifx\csname l@#1\endcsname\relax
\typeout{** WARNING: IEEEtran.bst: No hyphenation pattern has been}%
\typeout{** loaded for the language `#1'. Using the pattern for}%
\typeout{** the default language instead.}%
\else
\language=\csname l@#1\endcsname
\fi
#2}}
\providecommand{\BIBdecl}{\relax}
\BIBdecl

\bibitem{smith2009fragments}
L.~B. Smith, ``From fragments to geometric shape: Changes in visual object
  recognition between 18 and 24 months,'' \emph{Curr. Dir. Psychol. Sci.},
  vol.~18, no.~5, pp. 290--294, 2009.

\bibitem{ornkloo2007fitting}
H.~{\"O}rnkloo and C.~von Hofsten, ``Fitting objects into holes: On the
  development of spatial cognition skills.'' \emph{Dev. Psychol.}, vol.~43,
  no.~2, pp. 404--416, 2007.

\bibitem{higgins2017beta}
I.~Higgins, L.~Matthey, A.~Pal, C.~Burgess, X.~Glorot, M.~Botvinick,
  S.~Mohamed, and A.~Lerchner, ``beta-{VAE}: Learning basic visual concepts
  with a constrained variational framework.'' in \emph{Proc. Int. Conf. Learn.
  Representations}, 2017.

\bibitem{chen2016infogan}
X.~Chen, Y.~Duan, R.~Houthooft, J.~Schulman, I.~Sutskever, and P.~Abbeel,
  ``Info{GAN}: Interpretable representation learning by information maximizing
  generative adversarial nets,'' in \emph{Proc. Neural Inf. Process. Syst.},
  2016, pp. 2180--2188.

\bibitem{burgess2019monet}
C.~P. Burgess, L.~Matthey, N.~Watters, R.~Kabra, I.~Higgins, M.~Botvinick, and
  A.~Lerchner, ``{MONet}: Unsupervised scene decomposition and
  representation,'' \emph{arXiv preprint arXiv:1903.00450}, 2019.

\bibitem{greff2019multi}
K.~Greff, R.~L. Kaufman, R.~Kabra, N.~Watters, C.~Burgess, D.~Zoran,
  L.~Matthey, M.~Botvinick, and A.~Lerchner, ``Multi-object representation
  learning with iterative variational inference,'' \emph{arXiv preprint
  arXiv:1903.00450}, 2019.

\bibitem{locatello2019challenging}
F.~Locatello, S.~Bauer, M.~Lucic, G.~Raetsch, S.~Gelly, B.~Sch{\"o}lkopf, and
  O.~Bachem, ``Challenging common assumptions in the unsupervised learning of
  disentangled representations,'' in \emph{Proc. 36th Int. Conf. Mach. Learn.},
  vol.~97, 2019, pp. 4114--4124.

\bibitem{otsu1986recognition}
N.~Otsu, ``Recognition of shape and transformation: An invariant-theoretical
  foundation,'' in \emph{Proc. 1st Int. Symp. Sci. Form}, 1986, pp. 413--420.

\bibitem{miao2007learning}
X.~Miao and R.~P. Rao, ``Learning the {Lie} groups of visual invariance,''
  \emph{Neural Comput.}, vol.~19, no.~10, pp. 2665--2693, 2007.

\bibitem{olshausen2007bilinear}
B.~A. Olshausen, C.~Cadieu, J.~Culpepper, and D.~K. Warland, ``Bilinear models
  of natural images,'' in \emph{Proc. SPIE 6492, Human Vision Electron. Imag.
  XII}, vol. 6492, 2007, pp. 67--76.

\bibitem{DBLP:journals/corr/abs-1001-1027}
J.~Sohl{-}Dickstein, J.~C. Wang, and B.~A. Olshausen, ``An unsupervised
  algorithm for learning {Lie} group transformations,'' \emph{arXiv preprint
  arXiv:1001.1027}, 2010.

\bibitem{memisevic2010learning}
R.~Memisevic and G.~E. Hinton, ``Learning to represent spatial transformations
  with factored higher-order boltzmann machines,'' \emph{Neural Comput.},
  vol.~22, no.~6, pp. 1473--1492, 2010.

\bibitem{chau2020disentangling}
H.~Y. Chau, F.~Qiu, Y.~Chen, and B.~Olshausen, ``Disentangling images with lie
  group transformations and sparse coding,'' \emph{arXiv preprint
  arXiv:2012.12071}, 2020.

\bibitem{takada2021unsupervised}
T.~Takada, Y.~Ohmura, and Y.~Kuniyoshi, ``Unsupervised learning of
  shape-invariant lie group transformer by embedding ordinary differential
  equation,'' in \emph{Proc. 2021 IEEE Int. Conf. Develop. Learn.}\hskip 1em
  plus 0.5em minus 0.4em\relax IEEE, 2021, pp. 1--6.

\bibitem{chen2018neural}
T.~Q. Chen, Y.~Rubanova, J.~Bettencourt, and D.~K. Duvenaud, ``Neural ordinary
  differential equations,'' in \emph{Proc. Neural Inf. Process. Syst.}, 2018,
  pp. 6571--6583.

\end{thebibliography}


\end{document}